\documentclass{article}

\usepackage{microtype}
\usepackage{graphicx}
\usepackage{subfigure}
\usepackage{booktabs} 

\usepackage{hyperref}
\usepackage{xcolor}
\usepackage{ulem}

\usepackage[accepted]{icml2023}

\usepackage{amsmath}
\usepackage{amssymb}
\usepackage{mathtools}
\usepackage{amsthm}
\usepackage{multirow}
\usepackage{mathrsfs}

\usepackage[capitalize,noabbrev]{cleveref}

\theoremstyle{plain}

\theoremstyle{definition}

\theoremstyle{remark}

\usepackage[textsize=tiny]{todonotes}

\icmltitlerunning{Solving Trojan Detection Competitions with Linear Weight Classification}

\begin{document}

\twocolumn[
\icmltitle{Solving Trojan Detection Competitions with Linear Weight Classification}




\begin{icmlauthorlist}
\icmlauthor{Todd Huster}{xxx}
\icmlauthor{Peter Lin}{yyy}
\icmlauthor{Razvan Stefanescu}{yyy}
\icmlauthor{Emmanuel Ekwedike}{zzz}
\icmlauthor{Ritu Chadha}{yyy}
\end{icmlauthorlist}

\icmlaffiliation{yyy}{Peraton Labs, Basking Ridge, NJ, USA. All rights reserved.}
\icmlaffiliation{zzz}{Google}
\icmlaffiliation{xxx}{Lynx Capital Partners}

\icmlcorrespondingauthor{Razvan Stefanescu}{razvan.stefanescu@peratonlabs.com}

\icmlkeywords{Trojan Detection}

\vskip 0.3in
]



\printAffiliationsAndNotice{}  

\begin{abstract}
Neural networks can conceal malicious Trojan backdoors that allow a trigger to covertly change the model behavior. 
Detecting signs of these backdoors, particularly without access to any triggered data, is the subject of ongoing research and open challenges. 
In one common formulation of the problem, we are given a set of clean and poisoned models and need to predict whether a given test model is clean or poisoned. 
In this paper, we introduce a detector that works remarkably well across many of the existing datasets and domains. It is obtained by training a binary classifier on a large number of models' weights after performing a few different pre-processing steps including feature selection and standardization, reference model weights subtraction, and model alignment prior to detection. We evaluate this algorithm on a diverse set of Trojan detection benchmarks and domains and examine the cases where the approach is most and least effective.

\end{abstract}

\section{Introduction}
 Trojan backdoors are hidden modifications in neural network models that allow an attacker to alter the model’s behavior in response to a specific trigger, posing significant risks to AI systems. The vulnerability of neural networks to Trojan backdoors is well documented.  Techniques for inserting triggers vary from simple data poisoning \cite{gu2019badnets} to clean label attacks \cite{Turner2018CleanLabelBA, saha2019hidden, Liu2018TrojaningAO} to weight manipulation \cite{Liu2018TrojaningAO, Garg2020CanAW}. There have been several recent surveys covering backdoor attacks \cite{NeuralTrojansSurvey2020,li2021backdoor, wang22}.

A variety of techniques for detecting Trojan behavior have emerged in recent years. These include detecting anomalous samples during neural network training or inference \cite{ chou2020sentinet, gao2020strip, chen2018detecting}, attempting to recover the Trojan trigger via trigger inversion \cite{NeuralCleanse2019, guo2019tabor, wang2020practical, sun2020poisoned, shen2021backdoor, huster21}, functional analysis \cite{sikka20, Xu2019DetectingAT, Edraki2020OdysseyCA, Erichson2020NoiseresponseAF}, activation analysis \cite{Tang2019DemonIT}, and weight analysis \cite{fields21, mlds}. 

In computer vision, techniques like activation clustering \cite{chen2018detecting} detect abnormal neuron activations that correspond to backdoor triggers, while Neural Cleanse \cite{NeuralCleanse2019} reverse-engineers potential triggers by identifying small input modifications that flip model predictions. Fine-pruning \cite{liu2018fine} is used to prune rarely activated neurons associated with triggers, effectively neutralizing the backdoor. ABS scanning \cite{shen2021backdoor}, detects neurons that respond abnormally to synthetic perturbations, uncovering hidden backdoors. Spectral signature analysis \cite{tran2018spectral} is also employed to identify outliers in neuron activations caused by backdoor inputs . These approaches aim to uncover hidden visual patterns or anomalies that activate backdoors in image-based models.

In natural language processing (NLP), backdoors typically appear as specific words or phrases that trigger malicious behavior. Detection techniques include input perturbation, where small modifications to text inputs help reveal triggers, and anomaly detection in embeddings, which identifies outliers in word embeddings or hidden states that correlate with backdoor behavior. Early stopping, perplexity and BERT Embedding distance were proposed in \cite{wallace2020concealed} to mitigate and identify poison examples in the training dataset. Traditional defense strategies, relying on model fine-tuning and gradient calculations, are insufficient for Large Language Models due to their computational demands, so the proposed Chain-of-Scrutiny (CoS) method \cite{li2024chain} detects backdoor attacks by generating and scrutinizing detailed reasoning steps to identify inconsistencies with the final answer.

In this paper, we introduce a simple, scalable, and powerful method for detecting Trojan backdoors across different domains including computer vision and NLP using linear weight classification. We focus on a common formulation of the problem where a set of clean and poisoned deep neural network models is provided, and the task is to predict whether a given test model is clean or poisoned. The detector is obtained by training a linear classifier on a large number of models' weights after performing a few different pre-processing steps. We start first by applying tensor and weight selection strategies resembling the first step in a forward-stagewise regression approach \cite{hastie2009elements}. Normalization was particularly effective when combined with reference model subtraction. We also explored permutation-invariant representations of tensors, and found that sorting was highly effective in addressing the arbitrary permutations of hidden units in trained neural networks. Our method falls under the category of weight analysis detection, which does not require any prior knowledge of the trigger or model outputs and is applicable across multiple domains.

We evaluate our approach on several benchmarks including datasets from the Trojan Detection Challenge (TDC22)\cite{TDC2022} and the IARPA/NIST TrojAI program\cite{trojaiframework}. The Trojan Detection Challenge(TDC22), a NeurIPS 2022 competition, tasks participants with detecting and analyzing Trojan attacks on deep neural networks designed to evade detection. The IARPA/NIST TrojAI program is a long-running initiative that has developed over 16 challenges using this formulation, addressing the issue of adversaries inserting Trojan behaviors into AI models by compromising the training pipeline. The program focuses on identifying such Trojans, which can be activated by specific triggers in an AI's input, causing the model to produce incorrect responses.

We also trained both clean and poisoned models from scratch using the Fashion MNIST dataset for our experiments. This dataset was especially valuable in demonstrating the importance of sorting tensors before training the logistic regression detector for neural networks initialized with random weights.

The structure of the paper is as follows: Section 2 discusses weight analysis methods for detecting backdoor models, followed by Section 3, which introduces our proposed methodology. Section 4 explains the experimental setup, including the evaluation metrics and datasets. The experimental results are presented in Section 5 followed by conclusions.

\section{Weight Analysis}

Detection methods that depend solely on the model's parameters, without analysing the model behaviour on any input data belong to the category of weight analysis techniques. Without the need of running the model or having any knowledge of input and output semantics, weight analysis is relatively simple to apply to different problem domains.  There have been a variety of proposed methods for extracting features from model parameters such as outlier detection \cite{fields21} and unsupervised clustering\cite{mlds}. Others, including extreme values, statistical moments, and operator norms, are commonly used in Trojan detection challenges.

\subsection{Weight Norm Analysis}

One simple set of features for weight analysis is the norms of the linear operators within the neural network. Typical norms include the spectral norm, $L^1$ or $L^{\infty}$ norm, and the Frobenius norm. Intuitively, since a poisoned model must react strongly to a small trigger, one might expect the norms of the linear operators to be larger in poisoned models than clean ones. 

\begin{figure}[H]
  \centering
    \includegraphics[width=.5\textwidth]{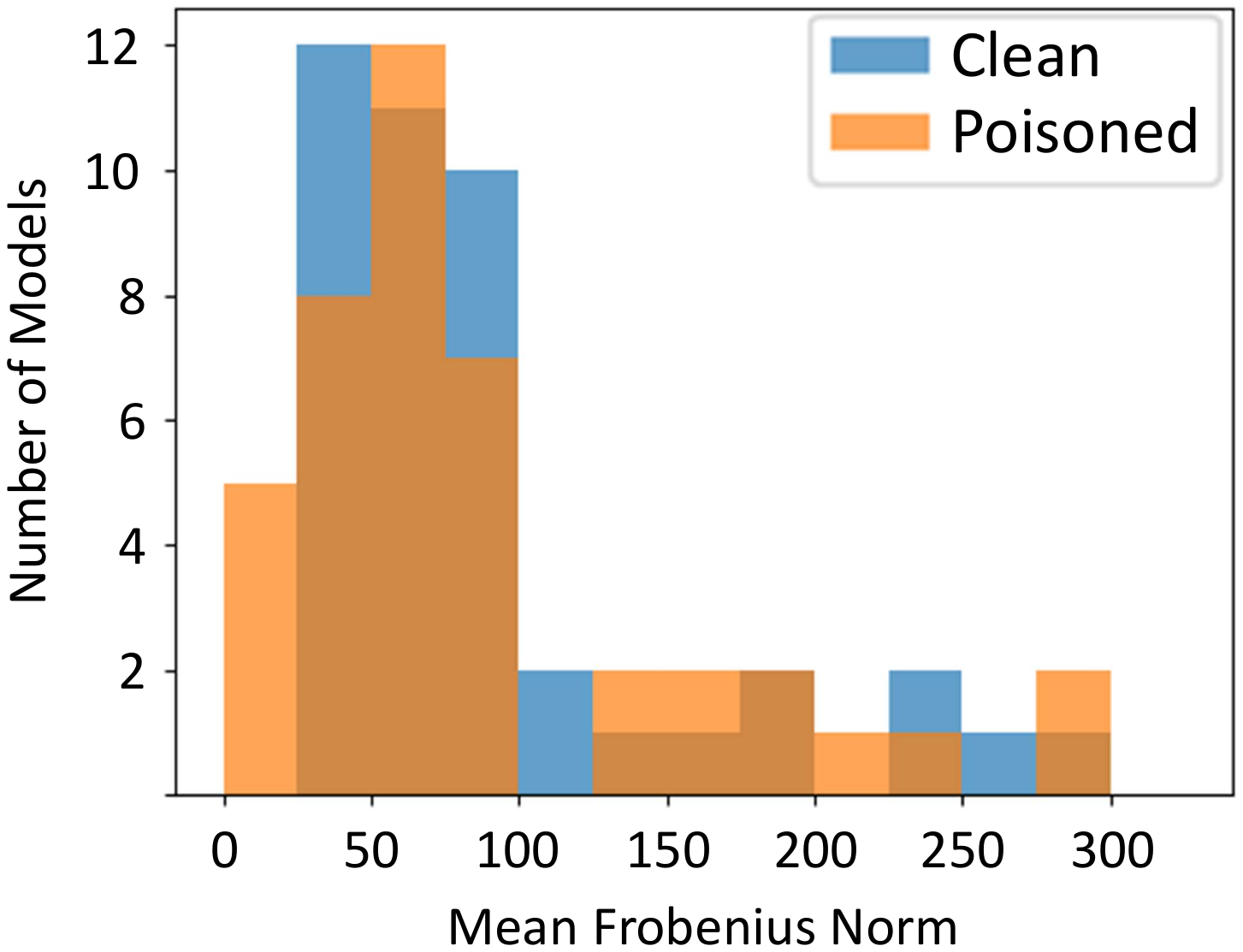} 
    \vspace{-11mm}
    \caption{Histogram of Average Frobenius Norms for ResNet models of the TrojAI NIST Round 11}\label{fig:fnorm} 
\end{figure}

Curiously, this intuitive justification for the use of weight norms turns out to be incorrect.  We plot the distribution of Frobenius norms of a set of clean and poisoned models from the September 2022 image classification challenge from the TrojAI program \cite{trojaiframework} in Figure \ref{fig:fnorm}. As is the case in most Trojan detection rounds, the distributions strongly overlap, and using this information as a detection statistic yields an equal error rate accuracy of 0.5.

That said, these same features clearly contain reliable information about whether or not the model is poisoned. 
We generated a feature vector for each model consisting of the Frobenius norms of each weight tensor, then trained a random forest on this set of features using default hyper-parameters from scikit-learn \cite{scikit-learn}. The model was able to predict whether held out models were clean or poisoned with 0.64 accuracy. While this is not an ideal classifier, there is clearly information about the poisoning in those values. 
We therefore have a set of somewhat discriminative features but not a strong justification for why these particular features should be effective.

\vspace{-7mm}

\section{Methodology}

Our proposed method starts from a simple question: are clean and poisoned model weights linearly separable? More precisely, if we treat each model parameter as a feature, can we find a hyperplane that reliably separates clean and poisoned models based on these features? As we will show in the remainder of this paper, the answer is often yes for the large majority of the backdoor problems. In the remainder of this section, we introduce our simple classification method and propose further optimizations that improve performance over the purely linear method.

\subsection{Linear Weight Classification}

We assume that we have multiple clean and poisoned models with the same architecture. 
We define two models as sharing the same architecture if they have identical layer types and topology, an equal number of parameter tensors, and parameter tensors of the same size. If two models have a few differently sized tensors, such as the final weight and bias tensors for models with different sets of classes, we exclude these tensors and still regard the models as having the same architecture. Most of the Trojan detection crounds provide multiple training models with shared architectures. 

Denote the $j$th tensor of model $i$ as $T^{i}_{j}$, which can have an arbitrary number of dimensions. Let $a$ denote the architecture of this model and $M_{a}$ denote the number of parameter tensors in this architecture. The flattened parameters of model $i$, $\mathcal{T}^{i}$, are thus denoted

\begin{align} \label{equ:tensors}
    \mathcal{T}^{i} = Flatten([T^{i}_{1}, ..., T^{i}_{M_a}]).
\end{align}

where the $Flatten$ operator deterministically arranges all input parameters into a single (one dimensional) vector. Let $y^i\in \{-1,1\}$ indicate that model $i$ is poisoned in case $y^i=1$ and clean if $y^i=-1$. 
Let $\mathcal{D}=\{(\mathcal{T}^{i}, y^i ) \mid i=1,...,N\}$ denote a dataset of N models sampled from a model generation process $\mathscr{D}$. 


Given a training dataset $\mathcal{D}_{train}$, the optimal separating hyperplane separates the two classes and maximizes the distance to the closest point from either class. A solution can be obtained by solving the following logistic regression optimization problem \cite{hastie2009elements}.

\begin{equation} \label{equ:lincls}
    \begin{aligned}
        \min_{W,b} \quad & - \frac{1}{N} \sum_{i=1}^{N} \left[ y^i \log(\hat{y}^i) + (1 - y^i) \log(1 - \hat{y}^i) \right] \\
        \text{where} \quad & \hat{y}^i = \frac{1}{1 + e^{-(W^T T^{i} + b)}}
    \end{aligned}
\end{equation}

\subsection{Feature Selection}

\subsubsection{Greedy Weight Selection}

Solving problem \ref{equ:lincls} can theoretically be accomplished by training an ordinary logistic regression classifier on $\mathcal{D}_{train}$. However, for many architectures $|\mathcal{T}^i|$ is very large, often tens or hundreds of millions (or even larger), which leads to practical issues optimizing and regularizing the classifier on $\mathcal{D}_{train}$.

The weights in a linear classifier represent how strongly each input feature influences the output (in this case, the detection score). To make reliable predictions, each weight must have a monotonic relationship with the detection score. This means that as the weight associated with a particular feature increases or decreases, the detection score must change in a consistent, predictable manner (either always increasing or always decreasing). We therefore look for weights that have a strong monotonic relationship with the labels $C$.

We consider the $k$th element of each model's vector $\mathcal{T}^{i}_{j}, j=1,..,M_a$ and construct array $\mathcal{T}^i[k]$.
We treat each $\mathcal{T}^i[k]$ as a detection statistic to predict the label $C$, train logistic regression models and compute the area under the receiver operating curve (ROC-AUC or simply AUC). Since a weight can have a positive or negative monotonic relationship with the class, we look for weights whose AUC score is as far away as possible from the uninformative value of $0.5$. We select the top $1000$ features, ordered by $\sigma_k$ criterion which we define as

\begin{align} \label{equ:auc}
    \sigma_k = \bigg| AUC\bigg( \mathcal{M}\big[ (\mathcal{T}^{i}[k],y^i) \mid i=1,...,N \big] \bigg) - 0.5 \bigg|,
\end{align}
where $\mathcal{M}$ is a logistic regression model trained on dataset $(\mathcal{T}^{i}[k],y^i) \mid i=1,...,N$. 

This gives us a manageable number of features that are well-suited for use in the linear classifier. We also considered using the correlation between $\mathcal{T}^i[k]$ and $y^i, i=1,..,N$ as the basis for feature selection, but we generally found the AUC score to be a slightly superior selection criterion. Weight selection was applied for all of our tested configurations.


\subsubsection{Tensor Selection}

In addition to the weight selection, we also selected layers that individually discriminate between clean and poisoned models. Additionally, some parameter tensors are significantly longer than others. For example, a model may have a normalization layer parameter with $64$ values, a fully connected layer with a million weights and a token embedding matrix with $20$ million weights. The normalization layer weights may actually be the most informative features, but the sheer size of the embedding matrix could cause it to dominate during the feature selection process. To minimize this imbalance, we identify the tensors with the most reliable features. 

We partition the training data into train and validation sub-splits. Then we train logistic regression models on each individual tensor. We assess the average AUC of these classifiers across multiple validation sub-splits and rank the tensors by how well they generalize. We then use the top $25$ tensors to select $\mathcal{T}^{i}$ and proceed with the remainder of our training process. When applied, tensor selection happened first followed by the weight selection. 

\subsection{Weight Normalization and Reference Model Subtraction}

One improvement we found over pure linear classification was normalizing the weights of each model. We experimented with a few normalization schemes, specifically normalizing each parameter tensor individually or normalizing the final flattened vector:

\begin{align} \label{equ:norm1}
    TensorNorm(T^{i}_{j}) = \frac{T^{i}_{j}}{StdDev(T^{i}_{j})}
\end{align}

\begin{align} \label{equ:norm2}
    ModelNorm(\mathcal{T}^{i}) = \frac{\mathcal{T}^{i}}{StdDev(\mathcal{T}^{i})}
\end{align}

In the above equations, we normalize based on the standard deviation. 
We also considered normalizing based on the $L^2$ or Frobenius norms. However, for neural network model weights, which are generally high dimensional with an expected value of zero, the $L^2$ norm and standard deviation are effectively identical down to a scalar multiple. 

Normalization was particularly effective when combined with reference model subtraction. In most of the TrojAI rounds, all models of a particular architecture are fine tuned from a common reference model. We subtract the reference tensors from each subject model tensors.


\begin{align} \label{equ:rsub}
    T^{i}_{j} \coloneq T^{i}_{j} - T^{ref}_{j}
\end{align}

By itself, subtracting a reference model from all subject models is equivalent to shifting the axes in feature space, and thus does not impact the linear separability of the classes. In practice, this yields a more robust classifier in cases where models have been fine tuned from a common source. 

\subsection{Permutation Invariance} 

It is well documented that there are many permutations of the same hidden units of a neural network that yield equivalent functions \cite{gitrebasin}. 
These permutations affect the ordering of rows and columns in weight matrices, input and output channel ordering in convolutional filters, dense layers, etc. 
A randomized training procedure is just as likely to arrive at one arrangement of these units as another.  This can introduce a lot of noise into our weight analysis pipeline since the models features end up being misaligned. 

To alleviate this issue, we propose exploring permutation invariant representations of tensors.  Such a representation should preserve as much information from the weights as possible, but should not be affected by arbitrary permutations of the hidden units. 

A simple and effective method for doing this appears to be tensor sorting. 
Given a tensor, we simply flatten it into a vector, then sort it. 
The elements of this sorted tensor can be thought of as a very high resolution set of quantiles of the tensor weights. 
While we lose some information about relative positions of weights, this gives us a permutation invariant representation of the tensor that maintains all weight values.

We explored other permutation invariant representations, including sorting on different combinations of dimensions, using singular vectors, and attempting to perform model alignment prior to detection. However, none of these methods reliably outperformed simple sorting. 

Permutation invariance seems to be especially critical in models trained starting from random weights initialization (scratch) (see Table \ref{tab:align} for illustration). 
In general, we found that tensor sorting in models that are fine-tuned from a pre-trained reference slightly degrades performance. It seems likely that in these cases, models are primed to represent concepts in a common fashion, so tensor sorting is destroying some useful information. However, tensor sorting dramatically outperforms on models trained from random weights initialization, where our base detector can struggle to perform significantly better than random guessing. Sorting is therefore more robust in this case.

\section{Experimental Setup}

\subsection{Evaluation Metrics}

We report area under the ROC curve (AUC) and cross entropy (CE), following the convention from the TrojAI program. AUC assesses how well a detector separates the clean and poisoned models and generally ranges from 0.5 (no separation) to 1.0 (perfect separation). The cross entropy between the predicted and true probability of poisoning captures both the \textit{accuracy} and \textit{calibration} of the detector. CE is therefore a strictly more difficult evaluation metric than AUC.  

\subsection{Data} 
We utilize the datasets from the IARPA Trojan AI (TrojAI) program and the 2022 Trojan Detection Challenge (TDC22). We also train our own small scale dataset for additional ablation experiments to show importance of permutation invariant transformation when building our detector. The remainder of this section describes the datasets and our procedures for running experiments with these datasets.

\subsubsection{TrojAI Datasets}

The IARPA TrojAI program has produced a series of Trojan detection challenge rounds \cite{trojaiframework}. In most rounds there is a set of labeled training models consisting of clean and poisoned neural networks that all solve a particular learning task. Tasks include image classification, object detection, named entity recognition (NER), sentiment analysis (SA), question answering (QA), and policy induction / reinforcement learning (RL). Computer vision (CV) and natural language processing (NLP) models were initialized from standard pre-trained models (e.g., from torchvision and HugginFace) and fine-tuned on their target tasks, while the RL models were initialized using random weights.

Participants use the training models to build a Trojan detector that is evaluated on separate test and holdout partitions of models. 
The TrojAI program maintains a test server that evaluates submitted detectors on the test partition and report results to a public leaderboard \footnote{https://pages.nist.gov/trojai/}.

We focus our evaluation on the currently active rounds with at least 40 training models per architecture, namely rounds 10, 11, and 13-16. We also include analysis on round 9, a diverse NLP round, in which test and holdout partitions have been made public. The domain, task(s), and number of architectures (\#Arch) and training models (\#TM) are shown in Table \ref{tab:rnd}. Gym refers to the domain of reinforcement learning and simulated environments for RL training supporting tasks such as robotic control, navigation, classic games (e.g., Atari games), and physics simulations \cite{moosavi2018multivariate}.

\begin{table}[!htbp]
\centering
\caption{Round Descriptions}
\label{tab:rnd}
\vspace{2mm}
\resizebox{.98\linewidth}{!}{
\begin{tabular}{|c|c|c|c|c|}\hline
    Rnd & Domain & Task(s) & \#Arch & \#TM\\\hline\hline
    9 & NLP & NER, SA, QA & 9 & 210\\\hline
    10 & CV & object detection & 2 & 144\\\hline
    11 & CV & image recognition & 3 & 288\\\hline
    13 & CV & object detection & 3 & 121\\\hline
    14 & Gym & policy induction & 2 & 238\\\hline
    15 & NLP& QA & 3 & 120\\\hline
    16 & Gym & policy induction & 2 & 222\\\hline
\end{tabular}
}
\end{table}

We trained seven logistic regression models with different configurations and applied them to the TrojAI datasets.  
These include our base configuration, which performs the most reliably on TrojAI rounds and ablation configurations designed to examine the impact of a given configuration parameter. 
Configurations are listed in Table \ref{tab:conf}. 
For a given configuration, the only tunable parameter is the logistic regression regularization term. We use scikit-learn library, with the regularization parameter $P$. We search for $P$ over the interval $[10^{-4}, 10^4]$ with logarithmic spacing independently for each classifier. For each choice of $P$, we run $30$ iterations of cross validation with $10\%$ of training models randomly held out. We select the value of $P$ that minimizes the cross entropy on the held out models. Next we fix $P$ and train the classifier on the full training dataset. 

\begin{table}[!htbp]
\centering
\caption{Detector configurations}
\label{tab:conf}
\vspace{2mm}
\resizebox{.98\linewidth}{!}{
\begin{tabular}{|c|c|c|c|c|}\hline
    \multirow{2}{*}{Name} & Reference & Norm & Tensor & \multirow{2}{*}{Sorted} \\
     & Model & Method & Selection &\\\hline\hline
     Base & Y & Tensor & Y & N\\\hline
     A & \textbf{N} & Tensor & Y & N\\\hline
     B & Y & \textbf{Model} & Y & Y\\\hline
     C & Y & Tensor & \textbf{N} & N\\\hline
     D & Y & Tensor & Y & \textbf{Y}\\\hline
     E & \textbf{N} & Tensor & Y & \textbf{Y}\\\hline
     F & \textbf{N} & \textbf{None} & Y & \textbf{Y}\\\hline
\end{tabular}
}
\end{table}

\subsubsection{TDC22 dataset}

The 2022 Trojan Detection Challenge followed a similar convention to a TrojAI round, but was developed independently as the topic of a NeurIPS workshop \footnote{https://2022.trojandetection.ai/index}. There are four image classification tasks with a single architecture per task (MNIST/five-layer CNN, CIFAR-10/ResNet18, CIFAR-100/ResNet18, and GTSRB/ViT). In contrast to the TrojAI CV models, all TDC22 models were initialized from random weights. Models from training, validation, and test partitions have all been released but labels are only available for the training partition. The training partition consists of 250 models for each task, half of which are poisoned. Since we do not have labels for the validation or test partitions, we ran multiple trials with 10\% of models held out for evaluation and average the AUC. To avoid overfitting on the test data, we set our hyperparameter $P = 3$, a value that proved robust in other experiments in terms of AUC. 

\subsubsection{Fashion MNIST dataset}
We trained a set of 400 models from scratch on the Fashion MNIST dataset \cite{xiao2017online}. 
We used two different triggers, namely the checkerboard and watermark triggers from \cite{huster21}. 
These triggers have small $L^1$ and $L^{\infty}$ norms, respectively. 
We used two different architectures: a five-layer CNN and a three-layer fully connected network. Table \ref{tab:fmnist} shows the average clean accuracy and attack success rate (ASR) of the models. 

\begin{table}[!htbp]
\centering
\caption{Fashion MNIST model accuracy and ASR. The architectures trained with no triggers are clean models.}
\label{tab:fmnist}
\vspace{2mm}
\resizebox{.98\linewidth}{!}{
\begin{tabular}{|c|c|c|c|}\hline
    Arch & Trigger & Clean Acc. & ASR \\\hline\hline
    FC & None & 0.90 & -- \\\hline
    FC & Watermark & 0.90 & 1.0 \\\hline
    FC & Checkerboard & 0.90 & 1.0 \\\hline
    CNN & None & 0.93 & -- \\\hline
    CNN & Watermark & 0.93 & 1.0 \\\hline
    CNN & Checkerboard & 0.93 & 1.0 \\\hline
\end{tabular}
}
\end{table}


\section{Experimental Results}

\subsection{TrojAI Leaderboard Results}

We show leaderboard results from active TrojAI rounds in Table  \ref{tab:trojres}. 
We compare all our configurations with the best competing method on the leaderboard, labeled other best. 
It is important to note that, in general, competing methods are tailored to each round, while we run the same method on all rounds. Our base method is highly effective on all rounds except R13, and tops all other methods in several round metrics, with the top score(s) for each round in bold. 

Round 13 was a very challenging round in which the training and test datasets have different distributions and triggers. Only one technique successfully achieved an AUC above 0.8 using a form of trigger inversion. Our detector failed on this round, which suggests our technique is sensitive to major changes in the distributions of the clean and poisoned models. 

\begin{table*}[!htbp]
\centering
\caption{TrojAI Leaderboard Results}
\label{tab:trojres}
\vspace{2mm}
\resizebox{.98\linewidth}{!}{

\begin{tabular}{|c|c|c|c|c|c|c|}\hline
    Config & R10 AUC / CE & R11 AUC / CE & R13 AUC / CE & R14 AUC / CE & R15 AUC / CE & R16 AUC / CE\\\hline\hline
    Base	&\textbf{0.96} / 0.26	&\textbf{0.99} / \textbf{0.10}	&0.52 / 1.14	&\textbf{1.00} / 0.08	&0.97 / \textbf{0.25}& 0.89 / 0.69\\\hline
A	&0.90 / 0.38	&0.97 / 0.23	&0.55 / 0.71	&\textbf{1.00} / 0.08	&0.90 / 0.46& 0.89 / 0.69\\\hline
B	&0.95 / 0.30	&\textbf{0.99} / 0.15	&0.54 / 1.10	&\textbf{1.00} / 0.10	&0.95 / 0.34& 0.83 / 0.74\\\hline
C	&0.94 / 0.31	&0.98 / 0.18	&0.51 / 0.78	&\textbf{1.00} / \textbf{0.05}	&0.91 / 0.38& 0.82 / 0.74\\\hline
D	&\textbf{0.96} / 0.28	&\textbf{0.99} / 0.14	&0.53 / 0.83	&\textbf{1.00} / \textbf{0.05}	&0.81 / 0.57& 0.87 / 0.65\\\hline
E	&0.92 / 0.39	&0.98 / 0.19	&0.54 / 0.72	&\textbf{1.00} / \textbf{0.05}	&0.91 / 0.41& 0.87 / 0.65\\\hline
F	&0.93 / 0.34	&0.93 / 0.36	&0.56 / 0.79	&0.99 / 0.09	&0.83 / 0.55& 0.81 / 0.79\\\hline\hline
Other Best &\textbf{0.96} / \textbf{0.17}	&0.96 / 0.29	&\textbf{0.93} / \textbf{0.28}	&\textbf{1.0} / 0.07	&\textbf{0.99} / 0.34& \textbf{1.0} / \textbf{0.06}\\\hline
\end{tabular}
}
\end{table*}

\subsection{TrojAI Round 9 Results}
\begin{figure}[H]
  \centering
    \includegraphics[width=.5\textwidth]{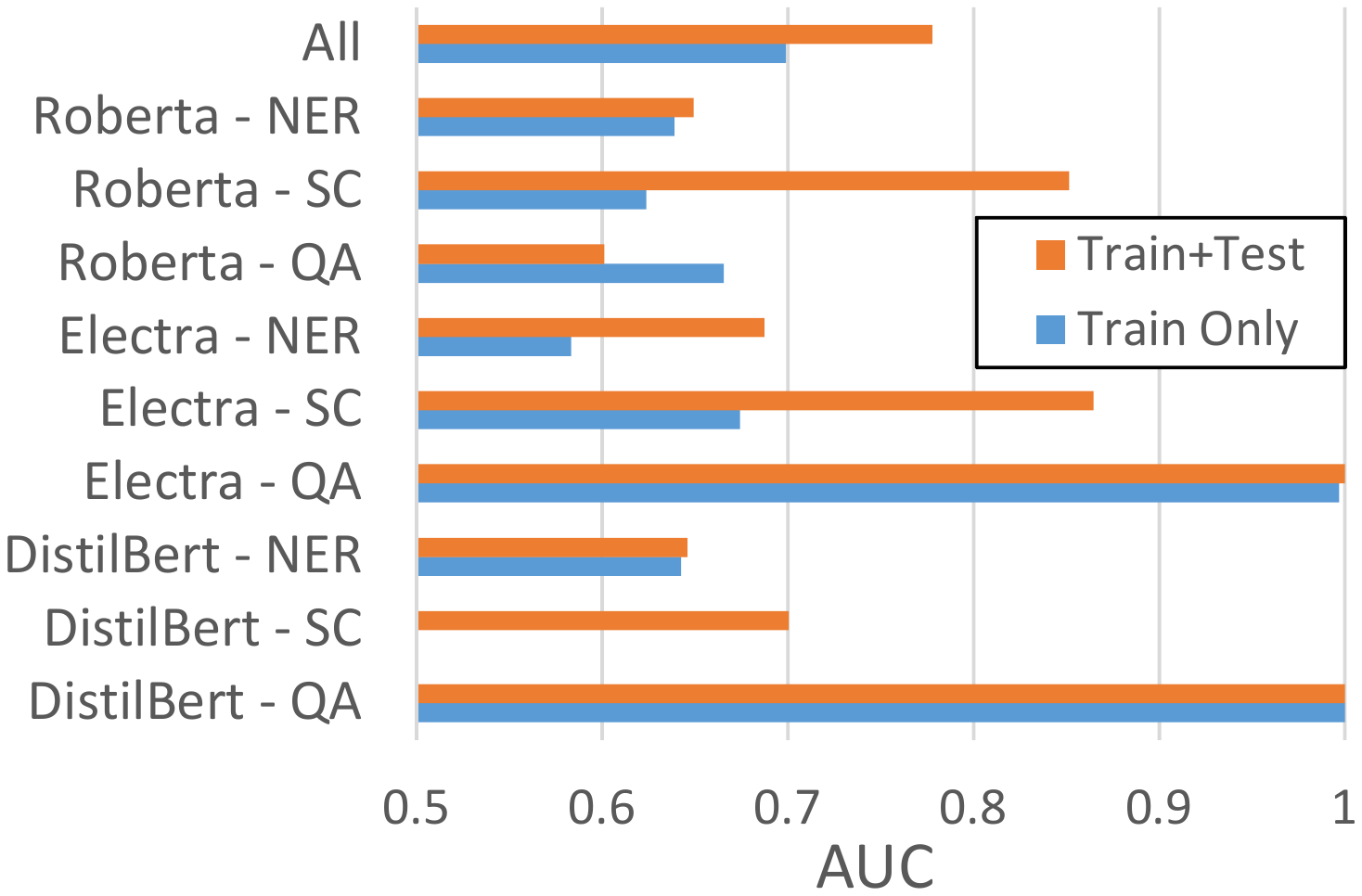} 
    \caption{Round 9 detection performance by architecture and task}\label{fig:r9} 
\end{figure}

TrojAI Round 9 featured three different tasks (NER, SC, QA) with three different architectures per task (RoBERTa, DistilBERT, and Electra). While there were 210 training models total, we needed to train a separate classifier for each of the nine task and architecture combinations, leaving only approximately $23$ training models per classifier, which proved challenging for our method. 
Some of these classifiers were still effective, but many had relatively poor performance. 
The competition organizers released the test and holdout partitions, both of which are three times larger than the training partition. 
To separate the impact of the small training set from the intrinsic difficulty of the round, we added models from the test partition into the training process and evaluated on the holdout partition. Figure \ref{fig:r9} shows the results. 
With the larger training set, all our classifiers are able to achieve an AUC of 0.6 or higher, with an average of 0.77. 
For the hardest task (QA), our detector performed better on less complex models such as  Electra and DistilBert than Roberta. This corroborates a previously reported conjecture that a model's spare capacity can be used to effectively hide a trigger, while adding a trigger to a model at capacity may leave more obvious signs of tampering that can be identified with weight analysis. 
\subsection{Features Alignment}
In this section, we focus on the permutation invariant strategy targeting TDC22 dataset, TrojAI round 11 and Fashion MNIST dataset. Table \ref{tab:align} provides the results of our base and D configurations with and without sorted tensors.
Without sorting, our method performs poorly on the TDC22 dataset, while it is very effective with sorting. 
As noted above, a key difference between TDC22 and most TrojAI rounds (including round 11) is that the TDC22 models are initialized with random weights instead of being fine-tuned from a pre-trained model. 
This difference may have an outsized impact on the linear separability of the models. 

\begin{table}[!htbp]
\centering
\caption{TDC22 and Round 11 results by architecture with (Config D) and without (Base) sorting}
\label{tab:align}
\vspace{2mm}
\resizebox{.98\linewidth}{!}{
\begin{tabular}{|c|c|c|c|c|}\hline
    \multirow{2}{*}{Source} & \multirow{2}{*}{Arch} & Task & Base & Config D \\
     &  & Dataset & AUC & AUC \\\hline\hline
    TDC22 & CNN & MNIST & 0.54 & 0.87 \\\hline
    TDC22 & ResNet & CIFAR10 & 0.71 & 1.0 \\\hline
    TDC22 & ResNet & CIFAR100 & 1.0 & 1.0 \\\hline
    TDC22 & ViT & GTSRB & 0.44 & 0.90 \\\hline
    R11 & MobileNet & CityScapes & 0.96 & 0.94 \\\hline
    R11 & ResNet & CityScapes & 1.0 & 0.80 \\\hline
    R11 & ViT & CityScapes & 0.98 & 0.81 \\\hline

\end{tabular}
}
\end{table}

\begin{figure}[H]
  \centering
    \includegraphics[width=.5\textwidth]{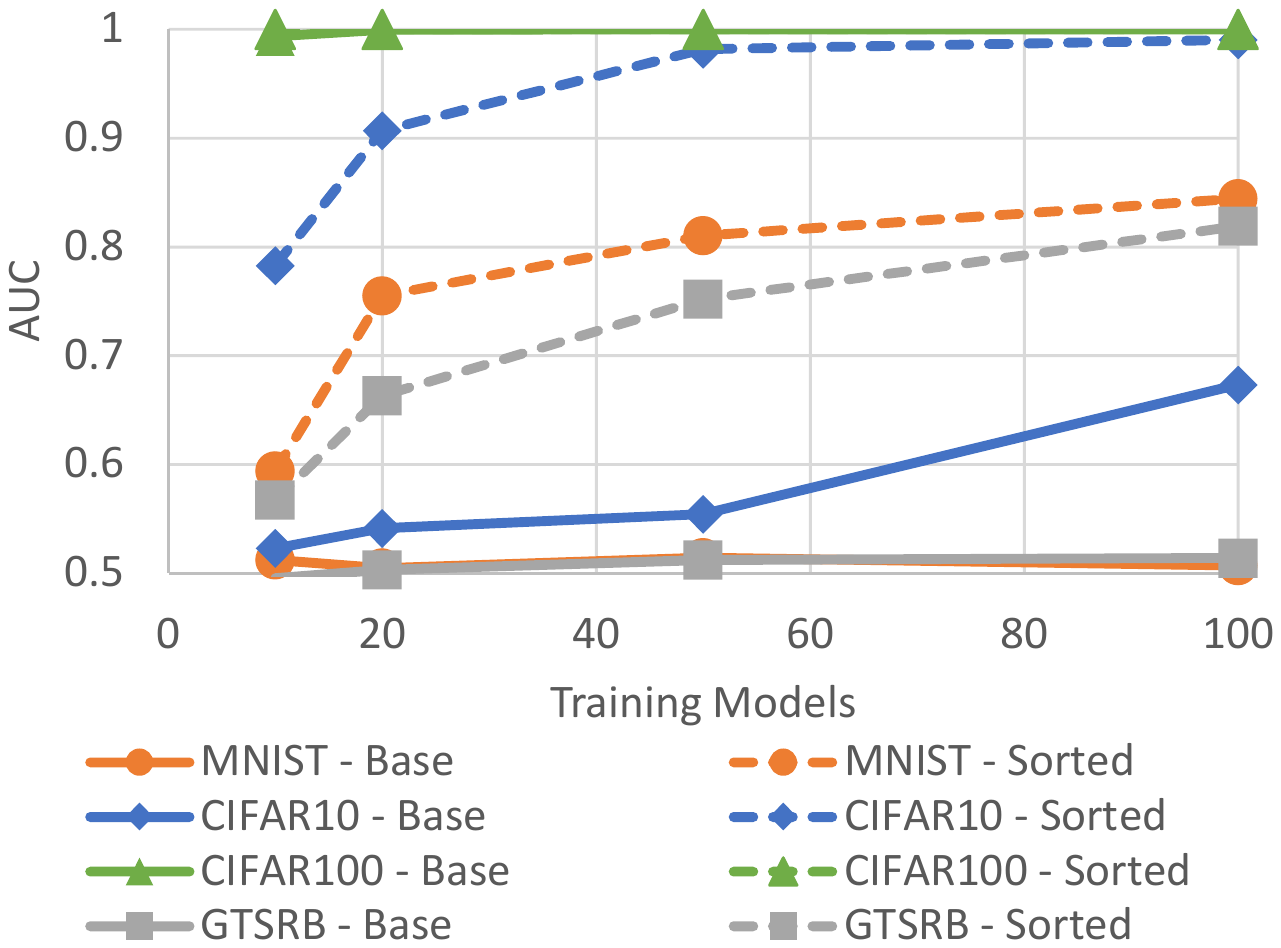} 
    \vspace{-11mm}
    \caption{TDC22 AUC vs. number of training models}\label{fig:tdc} 
\end{figure}

Figure \ref{fig:tdc} shows the impact of permutation invariant transformation and training set size on Trojan detection performance. 
While some of the architectures can be detected effectively with just a few training models, our method generally needs about $50$ training models to work effectively. 

\vspace{-7mm}

\begin{figure}[H]
  \centering
    \includegraphics[width=.5\textwidth]{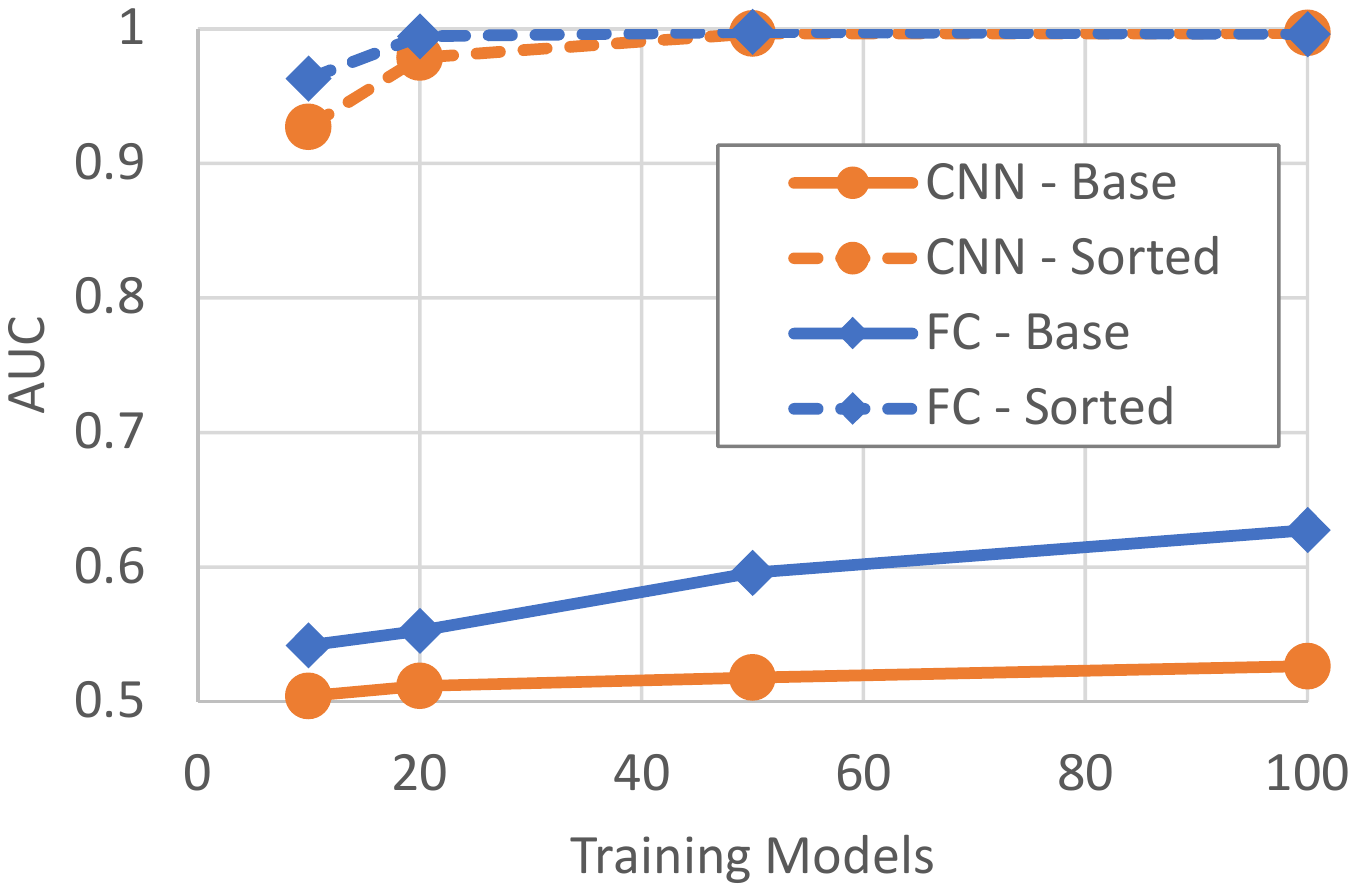} 
    \vspace{-11mm}
    \caption{Fashion MNIST AUC vs. number of training models}\label{fig:fmnist} 
\end{figure}

Our Fashion MNIST dataset enables us to explore two critical aspects of Trojan detection performance. 
First, since all models are initialized with random weights, we can explore the impact of model alignment on a different dataset than TDC22. 
Figure \ref{fig:fmnist} shows detection performance curves as a function of the number of training models. 
We show our detection performance with and without sorting on the FC and CNN architectures. 
With unsorted/unaligned base features, performance is very poor, especially for the higher capacity CNN models. 
With sorting, only $20$ training models are required to get performance above $0.95$ AUC for both architectures.

Secondly, we explore distribution shift by splitting the models by trigger type, namely the checkerboard and watermark. 
We split the data into two partitions - each had half the clean models and all poisoned models for one trigger type. 
We trained our method (both base config and config D with sorting) on each partition, then evaluated it on the other partition. 
Table \ref{tab:dshift} show the results. 
Even though the training and test partitions came from different distributions from trigger perspective, the aligned configuration D performed quite well, demonstrating that the method can be effective in non-IID settings. As expected, the unaligned base configuration performed poorly.

\vspace{-7mm}

\begin{table}[!htbp]
\centering
\caption{Fashion MNIST distribution shift results}
\label{tab:dshift}
\vspace{2mm}
\resizebox{.98\linewidth}{!}{
\begin{tabular}{|c|c|c|c|c|}\hline
    \multirow{2}{*}{Arch} & Train & Test & Base & Config D \\
     & Dataset & Dataset & AUC & AUC \\\hline\hline
    FC & WM+Clean & Checker+Clean & 0.48 & 0.83\\\hline
    CNN & WM+Clean & Checker+Clean & 0.45 & 0.97\\\hline
    FC & Checker+Clean & WM+Clean & 0.63 & 1.0\\\hline
    CNN & Checker+Clean & WM+Clean & 0.47 & 0.99\\\hline
\end{tabular}
}
\end{table}

\section{Conclusions}

In this paper, we have demonstrated that simple linear classifiers can be surprisingly effective at detecting Trojan backdoors in neural networks. Unlike many other Trojan detection methods, our method can be easily applied across diverse domains including computer vision, nlp and reinforcement learning and simulated environments, architectures, and triggers without extensive reworking. We have developed and demonstrated a set of simple modifications that incorporate additional sources of information when available (e.g., reference models), features selection and normalizations. We also introduced a permutation invariant sorting technique, which enables the method to succeed when presented with unaligned models. Our method demonstrated high performance across various benchmarks such as Trojan Detection Challenge (TDC22) and the IARPA/NIST TrojAI program, indicating that clean and poisoned models can be reliably separated using linear methods, provided the weights are properly aligned.

However, our method is dependent on a sufficient amount of representative training models. The method is robust to some types of distribution shift, as exhibited in our Fashion MNIST experiments, but can be negatively affected, as evidenced by the TrojAI round $13$ results. A thorough exploration and mitigation of this shortcoming is a topic for future work. Furthermore, signs of poisoning may be fundamentally hard to find via weight analysis in models with significant excess capacity relative to their trained task. An interesting unexplored strategy for defending against poisoning would be to ensure that models do not have significantly more capacity than necessary to perform effectively on their tasks. This strategy would make poisoning more conspicuous and weight analysis more effective at detecting abnormalities.

\section*{Acknowledgments}

\noindent This effort was supported by the Intelligence Advanced Research Projects Agency (IARPA) under the contract W911NF20C0034. The content of this paper does not necessarily reflect the position or the policy of the Government, and no official endorsement should be inferred.

\bibliography{main}
\bibliographystyle{icml2023}


\end{document}